\documentclass[letterpaper, 10 pt, conference]{ieeeconf}
\IEEEoverridecommandlockouts
\usepackage{amsmath}
\usepackage{amssymb}
\usepackage{bm}
\usepackage{booktabs}
\usepackage{xcolor}
\usepackage{multirow}
\usepackage{graphicx} 
\usepackage{colortbl}
\usepackage{cuted}
\usepackage{capt-of}
\usepackage{threeparttable}
\usepackage{placeins}
\usepackage{float}
\usepackage{subfigure}
\usepackage{subcaption}

\title{\LARGE \bf \underline{D}ual \underline{C}ovariance \underline{G}aussian \underline{S}platting SLAM: \\ Decoupling Rendering and Registration for Robust Real-Time Tracking}

\author{
    Edward Beng Wai Tan$^{1,*}$, Siew-Kei Lam$^{1}$%
    \thanks{$^{1}$College of Computing and Data Science, Nanyang Technological University, Singapore.}%
    \thanks{$^{*}$Corresponding author.}%
}

\begin{document}
\maketitle
\thispagestyle{empty}\pagestyle{empty}

\begin{strip}
    \vspace{-2.5\baselineskip}
    \centering
    \includegraphics[width=\textwidth]{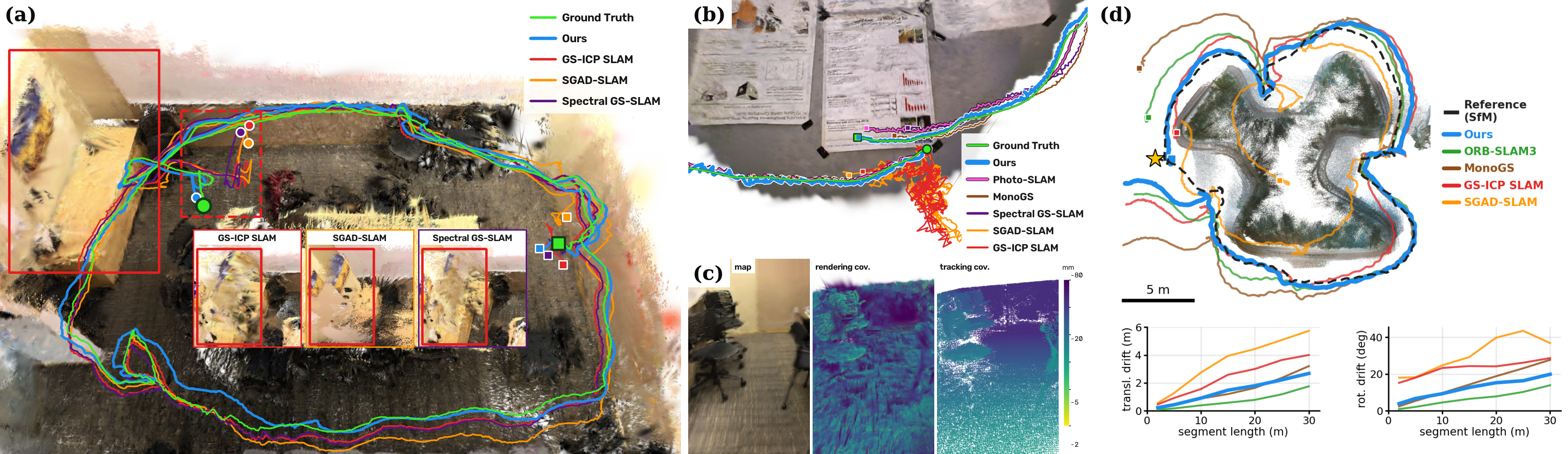}

    \captionof{figure}{Our method tracks robustly in various challenging conditions, such as indoor loops in (a) compared to other ICP methods, and in structure-less scenes rivaling MonoGS and Photo-SLAM in (b), and in outdoor scenes (d) shows the lowest drift among evaluated 3DGS-based methods. We do this by maintaining two covariances per Gaussian: one for rendering, and one for tracking, as shown in (c), colored by the size of the respective covariances.
    }
    \label{fig:hook}
\end{strip}
\thispagestyle{empty}
\pagestyle{empty}

\begin{abstract}
ICP-based 3D Gaussian Splatting (3DGS) SLAM tracks in real time by registering incoming frames against map Gaussians, using each primitive's covariance for both rendering and registration. These two uses place conflicting demands on one covariance. The mapper shapes it to minimize photometric error, often flattening it against surfaces, while robust registration typically benefits from measurement uncertainty. We propose a dual-covariance parameterization. Each Gaussian keeps a single mean but holds two covariances: a rendering covariance optimized by the mapper, and a tracking covariance derived from an RGB-D sensor noise model. We further use the tracking covariances as Gaussian anchors for image corners, providing constraints in directions where depth geometry is weak.
We evaluate on TUM RGB-D, ScanNet, Replica, and two outdoor sequences recorded with a RealSense D435i on wheeled and handheld platforms. We achieve robust tracking performance across multiple scenes and reduced odometry drift, while tracking at $\sim$ 60 FPS. 
\end{abstract}

\section{INTRODUCTION}
Recently, 3D Gaussian Splatting (3DGS) \cite{kerbl3DGaussianSplatting2023} has been incorporated into SLAM for photorealistic mapping. Initially, the pose was determined by dense photometric error \cite{matsukiGaussianSplattingSLAM2024, yanGSSLAMDenseVisual2024} due to the differentiable properties of 3DGS primitives. These methods were effective but proved to be computationally heavy, usually running at single digit FPS on desktop GPUs. 

To allow for online, real-time execution of 3DGS SLAM, various hybrid approaches were proposed, such as using sparse feature matching \cite{huangPhotoSLAMRealtimeSimultaneous2024} or the aligning point clouds \cite{haRGBDGSICPSLAM2025} using ICP. The ICP methods recognized that the explicit Gaussian parameterization compared to earlier neural implicit methods, could be used as both mapping primitives and as point clouds for alignment. 

In sharing the primitive, these methods also reused the Gaussian's covariance, optimized by the mapper for minimizing photometric error. This method proved to be sufficient for tracking in many scenes, yet \cite{tanSpectralGSSLAMObservabilityAware2026} demonstrated that it failed under specific adverse conditions, such as geometrically ill-conditioned scenes, due to lack of observability. We additionally find that the primitive's covariance itself does not encode sensor uncertainties accurately, due to its shared use with the mapper, often recording confidence in the incorrect direction with respect to the camera ray. 

To mitigate this, we propose a system where each Gaussian receives two covariances: one for tracking and one for mapping, while sharing the mean. This decoupling allows for the Gaussian to act as both a metric uncertainty model, and as a 3DGS primitive. We show that our method results in general tracking improvement across multiple datasets and camera types, reduced odometry drift, and good mapping quality, while maintaining real-time tracking performance. Finally, we evaluate the scene-level impact our method has on observability compared to other ICP-based 3DGS SLAM methods. To that end, we list our contributions:
\begin{itemize}
    \item We introduce a dual-covariance parameterization using 3D Gaussians for SLAM tracking, where one covariance stores the sensor uncertainty measurements, and the other is used for mapping.
    \item We propose a joint tracking strategy which uses both the sensor-derived covariances for ICP tracking, and stable landmark Gaussians as Kanade-Lucas-Tomasi (KLT) anchors to improve observability.
    \item We show that our method is robust to various challenging conditions with reduced odometry drift.
\end{itemize}

\section{RELATED WORK}
\textbf{Tracking in 3DGS SLAM} was initially based on photometric methods \cite{yanGSSLAMDenseVisual2024, matsukiGaussianSplattingSLAM2024, keethaSplaTAMSplatTrack2024} which used the rasterized Gaussian appearance to optimize for pose. These methods produced high quality maps but often had relatively low FPS, making them largely unsuitable for real-time tracking. Methods such as Photo-SLAM \cite{huangPhotoSLAMRealtimeSimultaneous2024} and GS-ICP SLAM \cite{haRGBDGSICPSLAM2025} introduced the use of classical tracking methods for high speed tracking. The ICP based methods were significantly faster compared to the earlier photometric tracking, but various works noted issues with geometry \cite{pakG2SICPSLAMGeometryaware2025}, observability \cite{tanSpectralGSSLAMObservabilityAware2026} and performance under RGB-D sensor noise \cite{haRGBDGSICPSLAM2025}. We argue that these issues are partly rooted in the use of the render-optimized Gaussian shape as the tracking covariance.

\textbf{Uncertainty weighted registration} proposed by G-ICP \cite{segalGeneralizedICP2009} allows for arbitrary covariances to weight the point-cloud registration. These covariances can be derived from RGB-D sensor uncertainty estimates \cite{nguyenModelingKinectSensor2012, khoshelhamAccuracyResolutionKinect2012}, yet the question remains on how the primitive should be modeled in the 3DGS case, when it is also optimized by photometric error.

\textbf{Observability and degeneracy} are long-standing problems in LiDAR SLAM for geometrically self-similar environments where some pose directions are left under-constrained, leading to degeneracy and tracking failure
\cite{tunaXICPLocalizabilityAwareLiDAR2024, yueLPICPGeneralLocalizabilityAware2025}.
This is typically diagnosed from the registration
Hessian \cite{zhangDegeneracyOptimizationbasedState2016}. Recently, \cite{tanSpectralGSSLAMObservabilityAware2026} showed that ICP-based 3DGS SLAM fails similarly in structure-less scenes, attributing this to the
mapper flattening the shared Gaussian covariances, and proposed detecting and
mitigating the resulting degeneracy. 

We distinguish two sources of this ill-conditioning. The first is \emph{weighting}: mapper flattened covariances assign spurious precision along surface normals, which a covariance reflecting actual sensor precision can correct. The second is \emph{geometric}: when the scene itself lacks structure in some direction,
as on a single plane, no reweighting of depth residuals can recover it, and
so additional information is required. Classical RGB-D tracking systems supply this by
combining geometric alignment with visual information, either as sparse feature
matches \cite{henryRGBDMappingUsing2014} or dense photometric error
\cite{whelanElasticFusion2016}. We therefore address the first problem with the sensor-derived tracking covariance, the second with KLT anchors to Gaussian primitives.

\section{METHOD}

\begin{figure*}[t]\centering
\includegraphics[width=\textwidth]{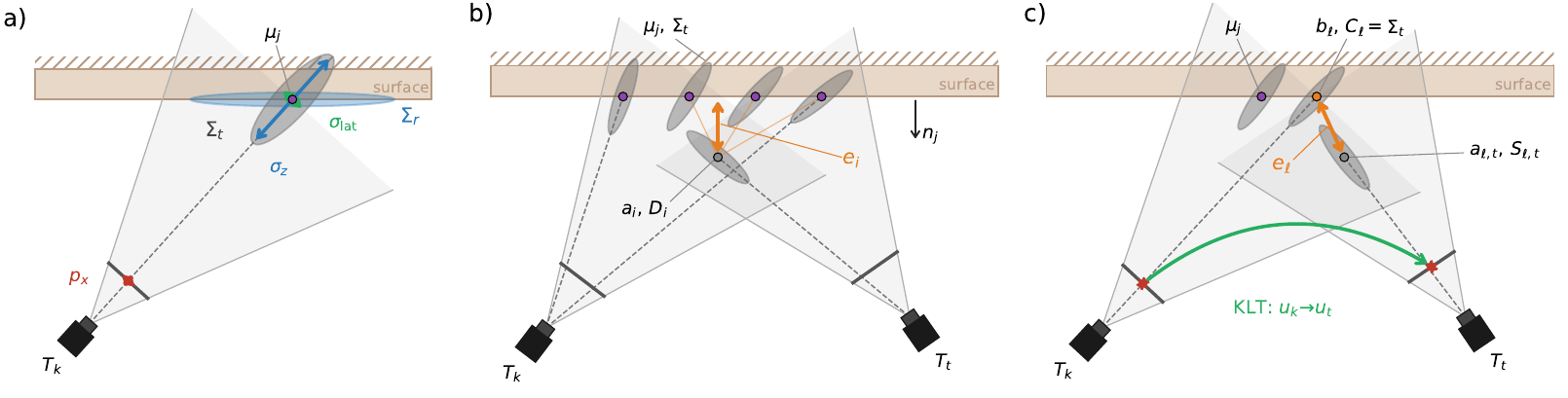}
\caption{\textbf{System Overview.} (a) We define each Gaussian to consist of a photometrically optimized covariance $\Sigma_r$ and a sensor measurement uncertainty covariance $\Sigma_t$. (b) The depth residuals are defined by the standard G-ICP cost with the source covariance $D_i$ defined as the observation's uncertainty. (c) The image residuals are defined as KLT corners anchored to nearby Gaussians, with uncertainty defined by the anchor's covariance.}\label{fig:mechanism}
\end{figure*}

\subsection{Preliminaries}
Let $a_i\in\mathbb{R}^3$ be the frame's points in camera coordinates and
$T=(R,t)\in\mathrm{SE}(3)$ the camera-to-world pose. Generalized-ICP~\cite{segalGeneralizedICP2009}
minimizes the residual
\begin{equation}
E(T)=\sum_i d_i^{\top}\big(C^B_{\pi(i)}+R\,C^A_iR^{\top}\big)^{-1}d_i,
\label{eq:gicp}
\end{equation}
where $d_i=b_{\pi(i)}-(Ra_i+t)$, and $C^A$ and $C^B$ are the metric covariances of the point positions and
$\pi$ is the nearest-neighbour association. 

3DGS primitives $\mathcal{G}_j = \{\mu_j, \Sigma_{r,j}, c_j, \sigma_j\}$ are used by GS-ICP SLAM~\cite{haRGBDGSICPSLAM2025} and related methods \cite{pakG2SICPSLAMGeometryaware2025, tanSpectralGSSLAMObservabilityAware2026, thirgoodFeatureSLAMFeatureenriched3D2026} by performing registration between the map Gaussians $b_j=\mu_j$, $C^B_j=\mathcal{Z}(\Sigma_{r,j})$ and
$C^A_i=\mathcal{Z}(Q_i)$, where $\Sigma_{r,j}$ is the rendering covariance,
$Q_i$ is the scatter of the $k$ nearest frame points and a regularizer
$\mathcal{Z}(\Sigma)=\Sigma/\lambda_2(\Sigma)$. Although $Q_i$ encodes the local geometric surface distribution, the map  covariances are also initialized from the KNN scatter. In practice, we find that the mapper's modification of the Gaussian mean $\mu_j$ sits well below depth characteristic noise for more than 97.5\% of cases when measured on the TUM RGB-D dataset. In contrast, prior work \cite{tanSpectralGSSLAMObservabilityAware2026} has found that the mapper tends to flatten
$\Sigma_{r,j}$ against the surface ($\lambda_3/\lambda_2\to0$), thus the target term assigns a small scale along the surface normal, despite the absence of corresponding depth precision in the 3DGS rendering model.

\subsection{Tracking covariance}
\label{subsec:trackcov}
We introduce a second covariance $\Sigma_{t,j}$ associated with each $\mathcal{G}_j$ alongside
$\Sigma_{r,j}$. The mapper optimises $\Sigma_{r,j}$ for rendering, and
$\Sigma_{t,j}$ is derived from the sensor uncertainty at the point of observation. A point
measured along the unit ray $v$ at range $z$ with lateral precision of $s$
pixels has approximate covariance, in m$^2$,
\begin{align}
& \mathcal{S}(v,z;s)=\sigma_\perp^2\big(I-vv^{\top}\big)+\sigma_\parallel^2\,vv^{\top},\\
& \sigma_\perp=\frac{zs}{f},\qquad
\sigma_\parallel^2=\sigma_z^2(z),
\end{align}
with $f$ the focal length and $\sigma_z(z)=A+Bz^2$ the structured-light range
law~\cite{nguyenModelingKinectSensor2012}. At each keyframe (or at spawn), the in-view Gaussians with optical center $o_k$ have their tracking covariance defined as
\begin{equation}
\Sigma_{t,j}=\mathcal{S}(u_j,z_j;p),
\qquad z_j=\lVert\mu_j-o_k\rVert,\quad u_j=(\mu_j-o_k)/z_j,
\label{eq:sigmat}
\end{equation}
with $p=0.5$\,px, so the anchor is the noise of the most recent observation of
the primitive.

\subsection{Depth residual}
\label{subsec:depthres}
On the observation side, we set the uncertainty to encode the measurement uncertainty plus the ambiguity of the ICP max. association radius, which lies in the local tangent plane, thus its covariance is defined as:
\begin{equation}
D_i=\mathcal{S}\big(v_i,\lVert a_i\rVert;p_d\big)+r_i^2\big(I-n_in_i^{\top}\big),
\qquad v_i=a_i/\lVert a_i\rVert,
\label{eq:rho}
\end{equation}
where $p_d$ is the pixel footprint of the depth sampling cell, $n_i$ is the
eigenvector of $Q_i$ with the smallest eigenvalue, and the association radius $r$
is the max. nearest-neighbour search distance. The depth
pair is therefore defined as $d_i=\mu_{\pi(i)}-Ta_i$ with
$\Omega_i=\Sigma_{t,\pi(i)}+R\,D_iR^{\top}$.

\subsection{Image residual}
\label{subsec:imgres}
Although the proposed second covariance models sensor uncertainty, it does not introduce constraints along directions lacking observability. Similar to classical RGB-D systems \cite{henryRGBDMappingUsing2014, whelanElasticFusion2016}, we utilize image information to supply the tangent-plane constraint that the depth observations lack. At each keyframe, a KLT corner back-projected to the world point $x^w$ near a primitive
$j$ is placed on that primitive's tangent plane,
\begin{equation}
\ell=x^w+n_jn_j^{\top}\big(\mu_j-x^w\big),
\end{equation}
where $n_j$ is the minimum-scale axis of $\Sigma_{r,j}$, roughly corresponding to the surface normal. The landmark keeps its tangential position from the image, takes its normal coordinate from the map,
and inherits the anchor's covariance $\Sigma_{t,j}$. In frame $f$ its KLT-tracked pixel is
back-projected at the measured depth $\hat z_\ell$ to $a_\ell$, with covariance
\begin{equation}
S_\ell=\mathcal{S}\big(v_\ell,\hat z_\ell;\sigma_m\big),
\qquad v_\ell=a_\ell/\lVert a_\ell\rVert,
\label{eq:slm}
\end{equation}
where $\sigma_m=2$\,px is the matcher's precision. This is \eqref{eq:rho}
without the association term, because the track fixes the correspondence. The
image pair is $e_\ell=\ell-Ta_\ell$ with
$\Omega_\ell=\Sigma_{t,j(\ell)}+R\,S_\ell R^{\top}$. In practice, as the matched set is slowly accumulated (and lost) over frames, this acts as a ``best effort" frame-to-map tracker, where landmarks are retained until KLT loses the corner due to sufficient viewpoint, illumination change, or other image artifacts. This is sufficient because its main purpose is to improve observability, rather than act as a primary tracker. In practice, our ablations show this additional residual contributes to a general increase in tracking robustness when applied \textit{in conjunction with} the tracking covariance.

\subsection{Estimation}
The pose is found by minimizing the term
\begin{equation}
E(T)=\sum_i d_i^{\top}\Omega_i^{-1}d_i+\sum_{\ell}\rho_\delta\big(e_\ell^{\top}\Omega_\ell^{-1}e_\ell\big)
\label{eq:cost}
\end{equation}
which represents the joint ICP and image residual with $\rho_\delta$ the Huber kernel, solved by Levenberg-Marquardt on
$\mathrm{SE}(3)$.

\section{EXPERIMENTS}\label{sec:res}
\subsection{Implementation and Experimental Setup}
\textbf{Datasets and Metrics.} We use TUM RGB-D \cite{sturmBenchmarkEvaluationRGBD2012}, ScanNet \cite{daiScanNetRichlyAnnotated3D2017}, Replica \cite{straubReplicaDatasetDigital2019} datasets to evaluate our work. We evaluate with Umeyama SE(3) aligned RMSE Absolute Trajectory Error (ATE) and Relative Pose Error (RPE) for measuring tracking accuracy, and map quality with PSNR, SSIM \cite{zhouwangImageQualityAssessment2004} and LPIPS \cite{zhangUnreasonableEffectivenessDeep2018}. We report all FPS as the average number of input frames processed over the entire sequence during SLAM tracking, excluding the input processing or post-hoc refinement time that some methods have.

\textbf{Implementation Details.} All of our evaluations were
performed on a computer equipped with an i9-13900KF CPU
and an RTX 4090 GPU. For reproduction, we evaluated GS-ICP SLAM \cite{haRGBDGSICPSLAM2025}, SGAD-SLAM \cite{Hu2026sgadslam}, Spectral GS-SLAM \cite{tanSpectralGSSLAMObservabilityAware2026}, Photo-SLAM \cite{huangPhotoSLAMRealtimeSimultaneous2024}, MonoGS \cite{matsukiGaussianSplattingSLAM2024}, SplaTAM \cite{keethaSplaTAMSplatTrack2024} and ORB-SLAM3 \cite{camposORBSLAM3AccurateOpenSource2021} using their publicly released implementations and default configurations.

\subsection{Tracking Performance}
We report our method's tracking results on the standard TUM sequences reported by most 3DGS SLAM methods in Table~\ref{tab:main_tum}. Compared to other `Coupled' methods, ours achieves the lowest ATE on average and lower than other methods on the fast-motion desk scene. Furthermore, ours has a significantly higher average FPS than the other two methods FeatureSLAM and SGAD-SLAM, which have comparable tracking performance to ours. Similar to ours, all the methods in the `Coupled' section are based on ICP tracking. The two `Decoupled' methods listed also have Bundle Adjustment (BA), while ours does not. We cap our method and GS-ICP SLAM at 30 FPS to evaluate ATE. We report the FPS, uncapped, as a separate run. G2S-ICP SLAM does not report their uncapped FPS.

\begin{table}[h]\centering
\caption{ATE [cm] $\downarrow$, on the standard TUM RGB-D scenes. Methods indicated with $\dagger$ are capped at 30 FPS. }\label{tab:main_tum}
\footnotesize\setlength{\tabcolsep}{5pt}
\resizebox{\columnwidth}{!}{%
\begin{tabular}{l l ccccc}
\toprule
\textbf{Type} & Method & fr1/desk & fr2/xyz & fr3/office & \emph{avg} & FPS $\uparrow$\\
\midrule
\multirow{2}{*}{Decoupled}
& ORB-SLAM3 \cite{camposORBSLAM3AccurateOpenSource2021} & \textbf{1.7} & 0.4 & 1.7 & \textbf{1.3} & \textbf{86.7} \\
& Photo-SLAM \cite{huangPhotoSLAMRealtimeSimultaneous2024} & 2.6 & \textbf{0.3} & \textbf{1.0} & \textbf{1.3} & 51.6 \\
\midrule
\multirow{5}{*}{Coupled}
& GS-ICP SLAM~\cite{haRGBDGSICPSLAM2025} $\dagger$ & 2.7 & 1.8 & 2.7 & 2.4 & \textbf{62.7} \\
& G2S-ICP SLAM \cite{pakG2SICPSLAMGeometryaware2025} $\dagger$ & 2.74 & {1.59} & 2.78 & 2.37 & - \\
& FeatureSLAM \cite{thirgoodFeatureSLAMFeatureenriched3D2026} & - &- &- & 2.05 & $\sim$5 \\
& SGAD-SLAM~\cite{Hu2026sgadslam}    & 2.2 & 1.7 & \textbf{2.0} & {2.0} & 20.4 \\

& \textbf{Ours} $\dagger$ & \textbf{1.94} & \textbf{1.55} & {2.23} & \textbf{1.91} & {59.1} \\
\bottomrule
\end{tabular}%
}
\end{table}

Additionally, we report extended results covering a more diverse scene set from TUM RGB-D in Table~\ref{tab:supp_tum}. These scenes cover a varying set of environments, including similar ones (\texttt{desk2}, \texttt{fr1/xyz}) to the ones in Table~\ref{tab:main_tum}, and more challenging ones (\texttt{room}, \texttt{360}), testing fast motion/rotation. We also test \texttt{s\_tex} as a reference scene containing both rich visual texture and geometry. As not all methods in Table~\ref{tab:main_tum} have released implementation, we report results on the available ones.

\begin{table}[t]\centering
\caption{Extended TUM RGB-D evaluation, ATE [cm] $\downarrow$.}\label{tab:supp_tum}
\footnotesize\setlength{\tabcolsep}{3pt}
\resizebox{\columnwidth}{!}{%
\begin{tabular}{l l cccccc c}
\toprule
\textbf{Type} & Method & desk2 & fr1/xyz & room & 360 & rpy & s\_tex & \emph{avg} \\
\midrule
\multirow{2}{*}{Decoupled}
& ORB-SLAM3~\cite{camposORBSLAM3AccurateOpenSource2021} & \textbf{2.23} & 1.06 & \textbf{6.47} & \textbf{20.93} & 0.43 & \textbf{0.98} &  \textbf{5.35}\\
& Photo-SLAM~\cite{huangPhotoSLAMRealtimeSimultaneous2024} &  2.77 & \textbf{1.00} & 6.95 & 21.6 & \textbf{0.31} & 0.99 &  5.60\\
\midrule
\multirow{4}{*}{Coupled}
& GS-ICP SLAM~\cite{haRGBDGSICPSLAM2025} & 13.34 & {1.43} & 18.08 & 42.30 & 2.51 & 2.14 & 13.30 \\
& Spectral GS-SLAM~\cite{tanSpectralGSSLAMObservabilityAware2026} & 12.18 & 1.44 & 17.35 & 29.24 & 2.34 & 3.80 & 11.06 \\
& SGAD-SLAM~\cite{Hu2026sgadslam}    & 7.21 & \textbf{1.40} & 27.87 & 80.14 & 15.19 & 2.13 & 22.33 \\
& Ours                & \textbf{3.83} & 1.47 & \textbf{16.23} & \textbf{17.13} & \textbf{2.24} & \textbf{1.82} & \textbf{7.12} \\
\bottomrule
\end{tabular}%
}
\end{table}

Our method is robust across challenging scenes containing a lack of visual features (\texttt{notexture}) and a lack of geometric structure (\texttt{nostructure}). We partitioned the results in Table~\ref{tab:degen} into methods that use depth for tracking only (top), and methods which have image-level signals (bottom). Ours maintains tracking robustly across both conditions, without any detection mechanism like \cite{tanSpectralGSSLAMObservabilityAware2026}. `X' denotes failed or partial tracking due to late initialization.

\begin{table}[h]\centering
\caption{TUM RGB-D no-structure/no-texture sequences, ATE [cm] $\downarrow$.}\label{tab:degen}
\footnotesize\setlength{\tabcolsep}{3pt}
\resizebox{\columnwidth}{!}{%
\begin{tabular}{l cccc}
\toprule
Method & nos\_tex\_near & nos\_tex\_far & s\_notex\_near & s\_notex\_far  \\
\midrule
GS-ICP SLAM~\cite{haRGBDGSICPSLAM2025} & 194.94 & 116.16 &  1.55 & 6.07  \\
SGAD-SLAM~\cite{Hu2026sgadslam}    & 195.08 & 116.17 & \textbf{1.35} & \textbf{1.94}  \\
\midrule
Photo-SLAM~\cite{huangPhotoSLAMRealtimeSimultaneous2024} & 2.27 & \textbf{5.36} & X & X \\
ORB-SLAM3~\cite{camposORBSLAM3AccurateOpenSource2021} & 2.03 & 10.07 & X & X \\
Spectral GS-SLAM~\cite{tanSpectralGSSLAMObservabilityAware2026}    & 7.79  & 16.85 & \textbf{1.55} & \textbf{2.19}  \\
Ours                & \textbf{1.67} & 9.00 & 1.86 & 4.93  \\
\bottomrule
\end{tabular}%
}
\end{table}

Our results on a subset of ScanNet are reported in Table~\ref{tab:scn}, where we achieve stable tracking performance compared to other methods, performing on par with methods like ORB-SLAM3 with both BA and loop closure, and dense photometric methods like SplaTAM and Gaussian-SLAM. Multiple scenes, e.g. \texttt{0169} have opportunities for loop closure. For SGAD-SLAM, we reproduced the results from the open-sourced code, using the authors' default per-scene tuning of the depth truncation and KNN initialization.

\begin{table}[h]\centering
\caption{ScanNet evaluation, ATE [cm] $\downarrow$.}\label{tab:scn}
\footnotesize\setlength{\tabcolsep}{3pt}
\resizebox{\columnwidth}{!}{%
\begin{tabular}{l cccccc c}
\toprule
 Method & 0000 & 0059 & 0106 & 0169 & 0181 & 0207 & \emph{avg} \\
\midrule
ORB-SLAM3 \cite{camposORBSLAM3AccurateOpenSource2021} & \textbf{8.38} & \textbf{7.11} & \textbf{9.57} & \textbf{8.45} & 67.68 & 7.74 & 18.2 \\
 SplaTAM \cite{keethaSplaTAMSplatTrack2024}
 & 12.8 & 10.1 & 17.7 & 12.1 & \textbf{11.1} & \textbf{7.5} & \textbf{11.9} \\
 Gaussian-SLAM \cite{yugayGaussianSLAMPhotorealisticDense2024}
 & 24.8 & 8.6 & 11.3 & 14.6 & 18.7 & 14.4 & 15.4 \\
 \midrule
 SGAD-SLAM~\cite{Hu2026sgadslam} \emph{(reported)}
 & 11.9 & 6.4 & 5.3 & 8.5 & 10.3 & 4.7 & 7.9 \\
 \midrule
 GS-ICP SLAM~\cite{haRGBDGSICPSLAM2025}
 & 36.5 & 22.17 & 6.04 & 31.26 & 15.8 & 10.47 & 20.4 \\
 SGAD-SLAM~\cite{Hu2026sgadslam}  $\ddagger$
 & 177.8 & 55.1 & 11.1 & 19.1 & 19.6 & 9.3 & 48.7 \\
  Ours
 & \textbf{15.63} & \textbf{8.67} & \textbf{5.59} & \textbf{6.89} & \textbf{13.06} & \textbf{7.83} & \textbf{9.61} \\
\bottomrule
\end{tabular}%
}
\vspace{2pt}
{\footnotesize $\ddagger$ indicates reproduced results by running the official code.}
\end{table}

We report our results on Replica in Table~\ref{tab:rep}, a simulated indoor scene without depth sensor noise, a near ideal case for ICP. As our covariance model described in Section~\ref{subsec:trackcov} requires a small non-zero depth noise floor for numerical stability, and this additional noise unsurprisingly results in marginally worse performance than other ICP methods, though it still outperforms other non-ICP 3DGS methods.

\begin{table}[h]\centering
\caption{Replica evaluation, ATE [cm] $\downarrow$.}\label{tab:rep}
\footnotesize\setlength{\tabcolsep}{3pt}
\resizebox{\columnwidth}{!}{%
\begin{tabular}{l cccccccc c}
\toprule
Method & r0 & r1 & r2 & o0 & o1 & o2 & o3 & o4 & \emph{avg} \\
\midrule
ORB-SLAM3 \cite{camposORBSLAM3AccurateOpenSource2021} & 0.56 & 0.43 & 0.40 & 0.90 & 0.67 & 1.55 & 0.96 & 2.29 & 0.97\\
SplaTAM \cite{keethaSplaTAMSplatTrack2024} & 0.31 & 0.40 & 0.29 & 0.47 &  0.27 & 0.29 & 0.32 & 0.55 & \textbf{0.36} \\
GS-SLAM \cite{yanGSSLAMDenseVisual2024} & 0.48 & 0.53 & 0.34 & 0.52 & 0.41 & 0.59 & 0.46 & 0.70 & 0.50 \\
MonoGS \cite{matsukiGaussianSplattingSLAM2024} & 0.48 & 0.36 & 0.34 & 0.44 & 0.52 & 0.23 & 0.16 & 2.53 & 0.63 \\
\midrule
G2S-ICP SLAM~\cite{pakG2SICPSLAMGeometryaware2025} & 0.14 & 0.16 & 0.10 & 0.19 & 0.12 & 0.16 & 0.17 & 0.20 & \textbf{0.15} \\
FeatureSLAM~\cite{thirgoodFeatureSLAMFeatureenriched3D2026} & -& -& -& -& -& -& -& -& \textbf{0.15} \\
GS-ICP SLAM~\cite{haRGBDGSICPSLAM2025} $\ddagger$ & {0.14} & {0.16} & 0.11 & 0.18 & {0.12} & {0.17} & {0.18} & 0.20 & {0.16} \\
SGAD-SLAM~\cite{Hu2026sgadslam} $\ddagger$    & 0.15 & {0.15} & {0.09} & {0.16} & {0.12} & 0.93 & 0.27 & {0.19} & 0.26 \\
Ours                & 0.16 & 0.17 & 0.36 & {0.16} & 0.13 & 0.18 & 0.20 & 0.22 & 0.20 \\
\bottomrule
\end{tabular}%
}
\vspace{2pt}
{\footnotesize $\ddagger$ indicates reproduced results by running the official code.}
\end{table}

We report our method's relative pose error in Table~\ref{tab:supp_tum_rpe}, which has the lowest overall error among the ICP-based methods, with drift characteristics more similar to ORB-SLAM3 than to the other ICP methods in both relative translation and relative rotation error over a 1 second window. This highlights our method's robustness to drift, due to the dedicated covariance used for tracking rather than reusing the mapping covariance.

\begin{table}[h]\centering
\caption{Extended TUM RGB-D evaluation, relative pose error over 1\,s without
trajectory alignment: translation RMSE [cm] $\downarrow$ / rotation RMSE [deg] $\downarrow$.}\label{tab:supp_tum_rpe}
\footnotesize\setlength{\tabcolsep}{3pt}
\resizebox{\columnwidth}{!}{%
\begin{tabular}{l l cccccc c}
\toprule
\textbf{Type} & Method & desk2 & fr1/xyz & room & 360 & rpy & s\_tex & \emph{avg} \\
\midrule
\multirow{2}{*}{Decoupled}
& ORB-SLAM3~\cite{camposORBSLAM3AccurateOpenSource2021} & 3.59/2.04 & 1.72/1.15 & 5.32/2.13 & 18.12/4.27 & 0.36/0.34 & 1.03/0.63 & 5.02/1.76 \\
& Photo-SLAM~\cite{huangPhotoSLAMRealtimeSimultaneous2024} & \textbf{3.35}/\textbf{1.88} & \textbf{1.54}/\textbf{1.00} & \textbf{4.30}/\textbf{1.80} & \textbf{14.41}/\textbf{3.40} & \textbf{0.29}/\textbf{0.32} & \textbf{0.96}/\textbf{0.59} & \textbf{4.14}/\textbf{1.50} \\
\midrule
\multirow{3}{*}{Coupled}
& GS-ICP SLAM~\cite{haRGBDGSICPSLAM2025} & 9.64/6.36 & 2.51/1.78 & 7.66/3.94 & 41.99/8.55 & 1.34/0.74 & 2.39/1.00 & 10.92/3.73 \\
& SGAD-SLAM~\cite{Hu2026sgadslam} & 10.96/6.46 & \textbf{2.50}/1.74 & 13.23/4.37 & 62.94/18.68 & 5.17/1.25 & 1.32/1.00 & 16.02/5.58 \\
& Ours & \textbf{5.36}/\textbf{3.67} & 2.61/\textbf{1.54} & \textbf{5.74}/\textbf{2.94} & \textbf{8.67}/\textbf{3.67} & \textbf{0.84}/\textbf{0.50} & \textbf{1.11}/\textbf{0.89} & \textbf{4.05}/\textbf{2.20} \\
\bottomrule
\end{tabular}%
}
\end{table}

\begin{figure}[h]
    \centering
    \includegraphics[width=0.49\columnwidth]{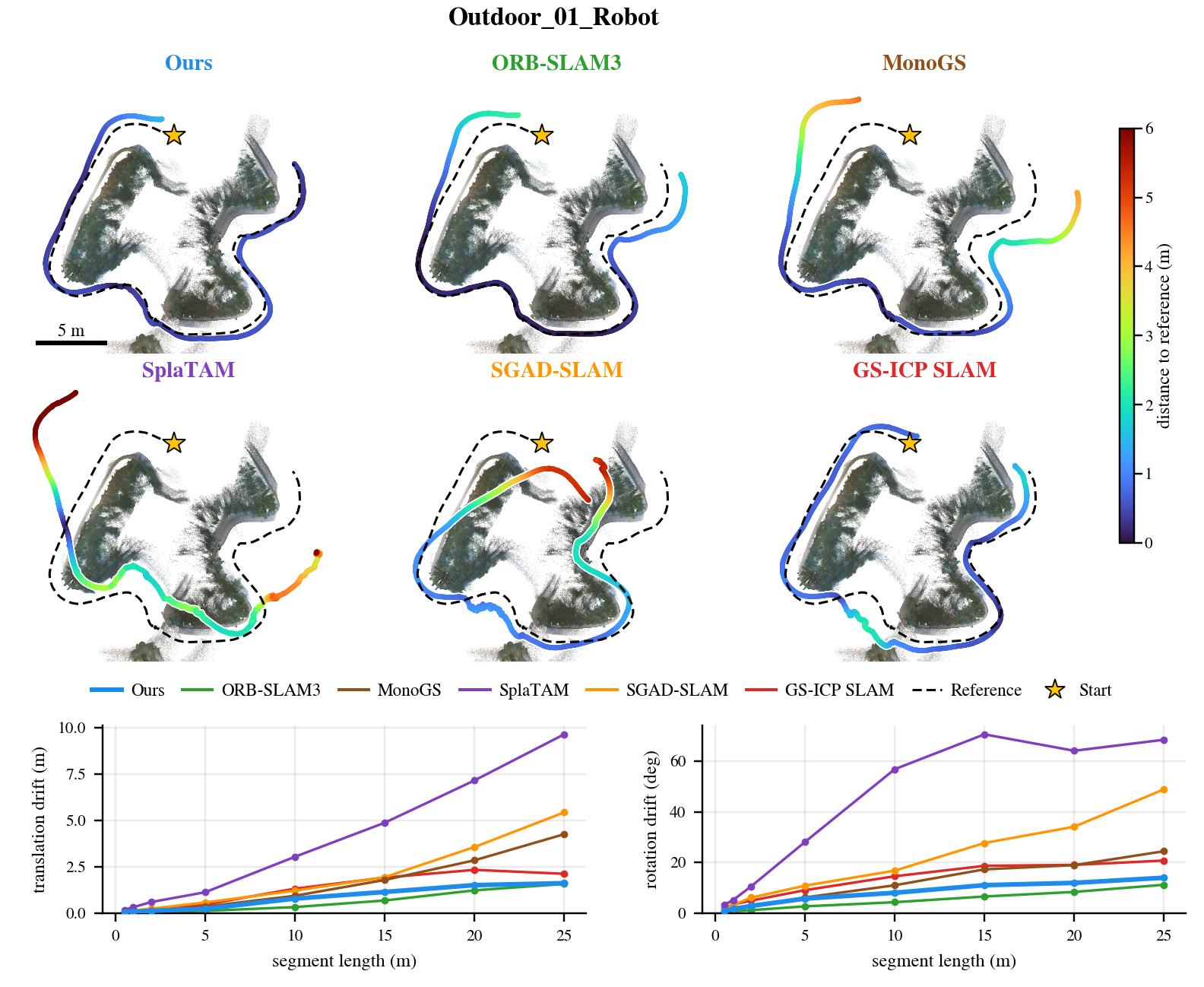}
    \hfill
    \includegraphics[width=0.49\columnwidth]{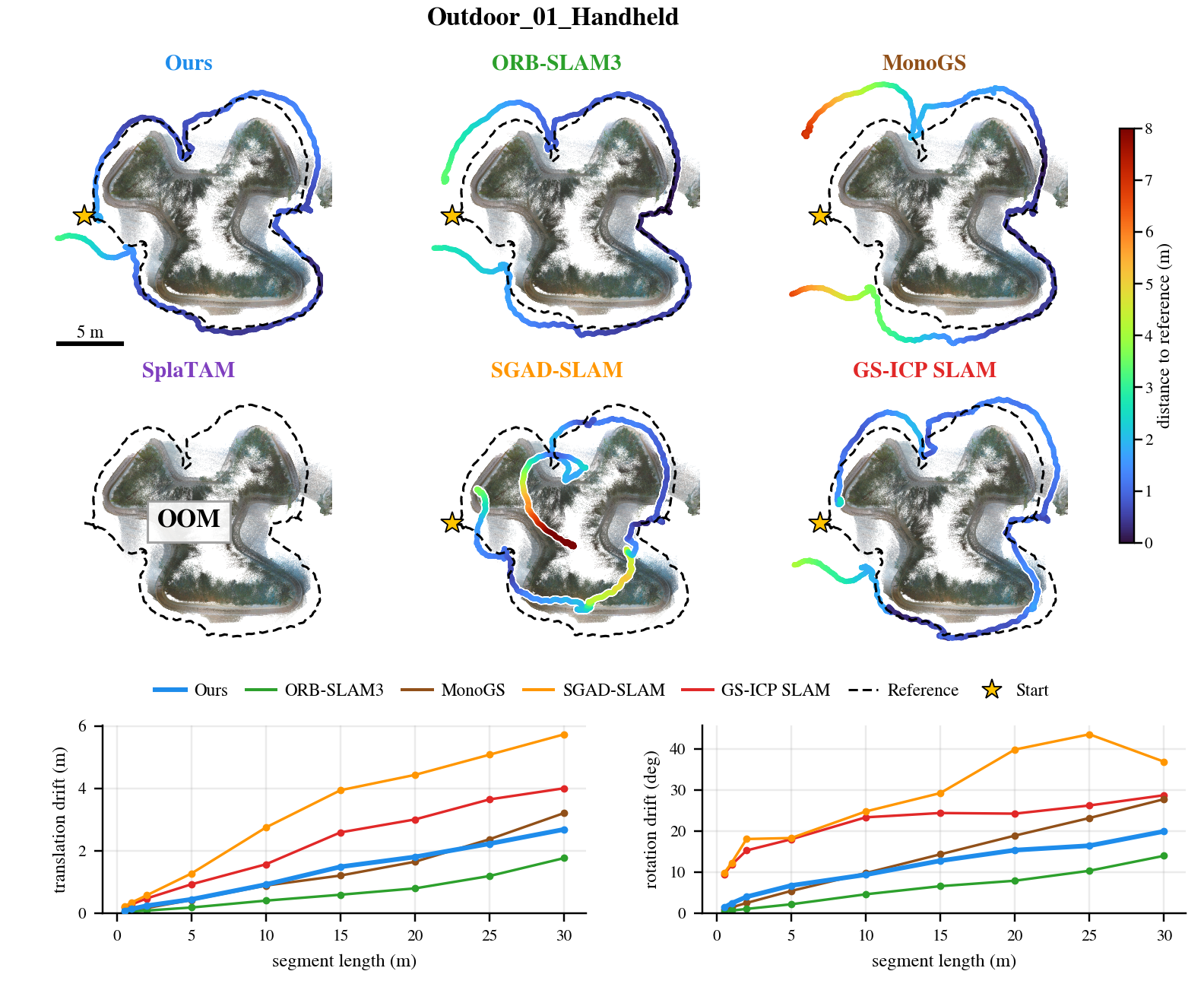}

    \caption{\textbf{Performance on outdoor scenes.}
    Ours has the lowest drift of all evaluated 3DGS SLAM methods in both
    the wheeled-platform (left) and handheld settings (right) on up to 70m+ length trajectories.}
    \label{fig:custom}
\end{figure}

\begin{table*}[t]\centering
\caption{Rendering performance of real-time 3DGS SLAM systems on TUM RGB-D}\label{tab:rendtum}
\footnotesize\setlength{\tabcolsep}{3pt}
\begin{tabular}{ll cccccccccc c}
\toprule
& Metric & desk & fr2/xyz & office & desk2 & fr1/xyz & room & 360 & rpy & s\_tex & nos\_tex & \emph{avg} \\
\midrule
\multirow{3}{*}{GS-ICP SLAM}
& PSNR\,$\uparrow$ & 17.99 & \underline{23.60} & 20.97 & 15.33 & 19.57 & 17.73 & \underline{18.05} & \underline{22.25} & 22.55 & 17.50 & 19.55 \\
& SSIM\,$\uparrow$ & 0.710 & \underline{0.836} & 0.764 & 0.674 & 0.739 & \underline{0.697} & \underline{0.727} & \underline{0.815} & 0.788 & 0.744 & 0.749 \\
& LPIPS\,$\downarrow$ & 0.293 & \underline{0.136} & 0.222 & 0.363 & 0.252 & \underline{0.316} & \underline{0.349} & \underline{0.170} & 0.227 & 0.516 & 0.284 \\
\midrule
\multirow{3}{*}{Spectral GS-SLAM}
& PSNR\,$\uparrow$ & 18.52 & 22.13 & 18.18 & 15.66 & 19.60 & 17.42 & 16.38 & 19.74 & 19.67 & 24.98 & 19.23 \\
& SSIM\,$\uparrow$ & 0.724 & 0.819 & 0.714 & 0.680 & 0.740 & 0.689 & 0.703 & 0.778 & 0.751 & \underline{0.849} & 0.745 \\
& LPIPS\,$\downarrow$ & 0.279 & 0.163 & 0.303 & 0.356 & 0.250 & 0.323 & 0.383 & 0.239 & 0.276 & 0.182 & 0.275 \\
\midrule
\multirow{3}{*}{Photo-SLAM}
& PSNR\,$\uparrow$ & \textbf{21.22} & 21.85 & \textbf{22.64} & \textbf{18.26} & \textbf{23.29} & \underline{18.14} & 15.44 & 20.53 & \textbf{25.47} & \textbf{26.95} & \underline{21.38} \\
& SSIM\,$\uparrow$ & \textbf{0.752} & 0.760 & \underline{0.774} & \underline{0.694} & \textbf{0.815} & 0.657 & 0.610 & 0.719 & \textbf{0.857} & \textbf{0.865} & \underline{0.750} \\
& LPIPS\,$\downarrow$ & \textbf{0.225} & 0.174 & \textbf{0.155} & \underline{0.325} & \textbf{0.167} & \underline{0.316} & 0.431 & 0.203 & \textbf{0.144} & \textbf{0.143} & \underline{0.228} \\
\midrule
\multirow{3}{*}{\textbf{Ours}}
& PSNR\,$\uparrow$ & \underline{18.87} & \textbf{24.52} & \underline{21.51} & \underline{18.04} & \underline{20.50} & \textbf{18.24} & \textbf{18.51} & \textbf{22.75} & \underline{24.47} & \underline{26.51} & \textbf{21.39} \\
& SSIM\,$\uparrow$ & \underline{0.731} & \textbf{0.859} & \textbf{0.781} & \textbf{0.725} & \underline{0.762} & \textbf{0.711} & \textbf{0.755} & \textbf{0.829} & \underline{0.827} & \textbf{0.865} & \textbf{0.784} \\
& LPIPS\,$\downarrow$ & \underline{0.267} & \textbf{0.114} & \underline{0.201} & \textbf{0.286} & \underline{0.216} & \textbf{0.300} & \textbf{0.321} & \textbf{0.151} & \underline{0.181} & \underline{0.157} & \textbf{0.219} \\
\bottomrule
\end{tabular}
\end{table*}

To evaluate our method's robustness to drift over long sequences, we recorded two sequences in an outdoor scene pictured in Figure~\ref{fig:custom}. These sequences were recorded on a wheeled platform and handheld respectively, using a D435i RGB-D camera. The pseudo-GT trajectories were generated offline using COLMAP \cite{schoenberger2016sfm} with scale from the RGB-D sensor. We run our method, with the other ICP-based baselines (GS-ICP SLAM, SGAD-SLAM), dense photometric methods (SplaTAM, MonoGS), and ORB-SLAM3 as reference. Despite having no bundle adjustment, our method reports the lowest drift of all evaluated 3DGS SLAM methods outperformed only by ORB-SLAM3, while simultaneously constructing a 3DGS map in real-time. We attempted a few other methods (e.g., Photo-SLAM) but there was insufficient memory to complete the sequence.

\subsection{Mapping Performance}
\begin{figure}[h]\centering
\includegraphics[width=\columnwidth]{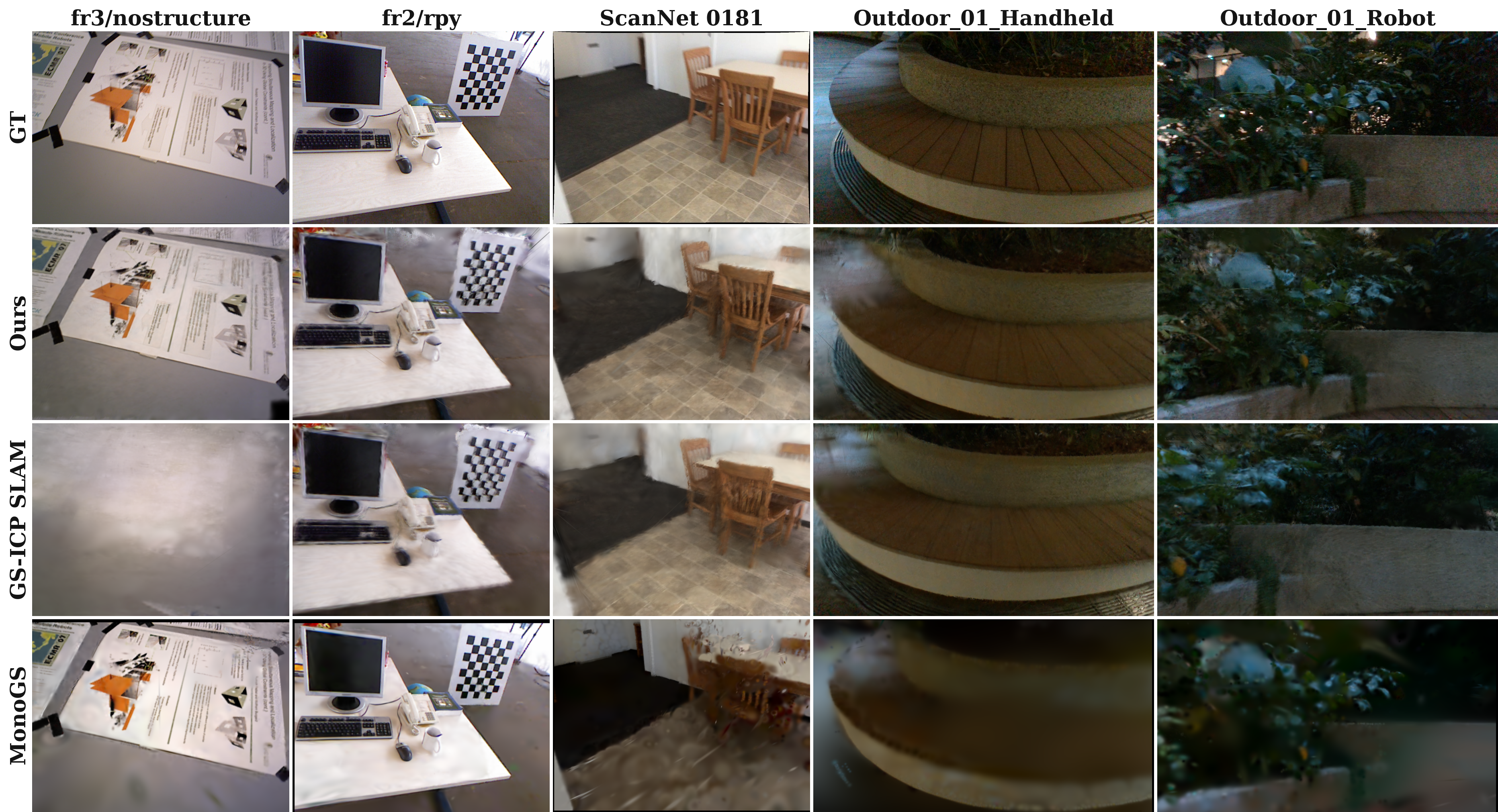}
\caption{\textbf{Qualitative render quality} of our method compared to some baselines. We render scenes from TUM RGB-D, ScanNet, and from our custom sequences.}\label{fig:qual_map}
\end{figure}

We report mapping performance by scene in Table~\ref{tab:rendtum}. Our method performs similarly to Photo-SLAM in PSNR and has the best SSIM and LPIPS of all real-time methods evaluated, in part due to the robust tracking, which allows for the mapping to take place from accurate viewpoints. We also provide qualitative renders of some of the methods in Figure~\ref{fig:qual_map}.

\subsection{Observability analysis} 
\begin{figure}[h]\centering
\includegraphics[width=\columnwidth]{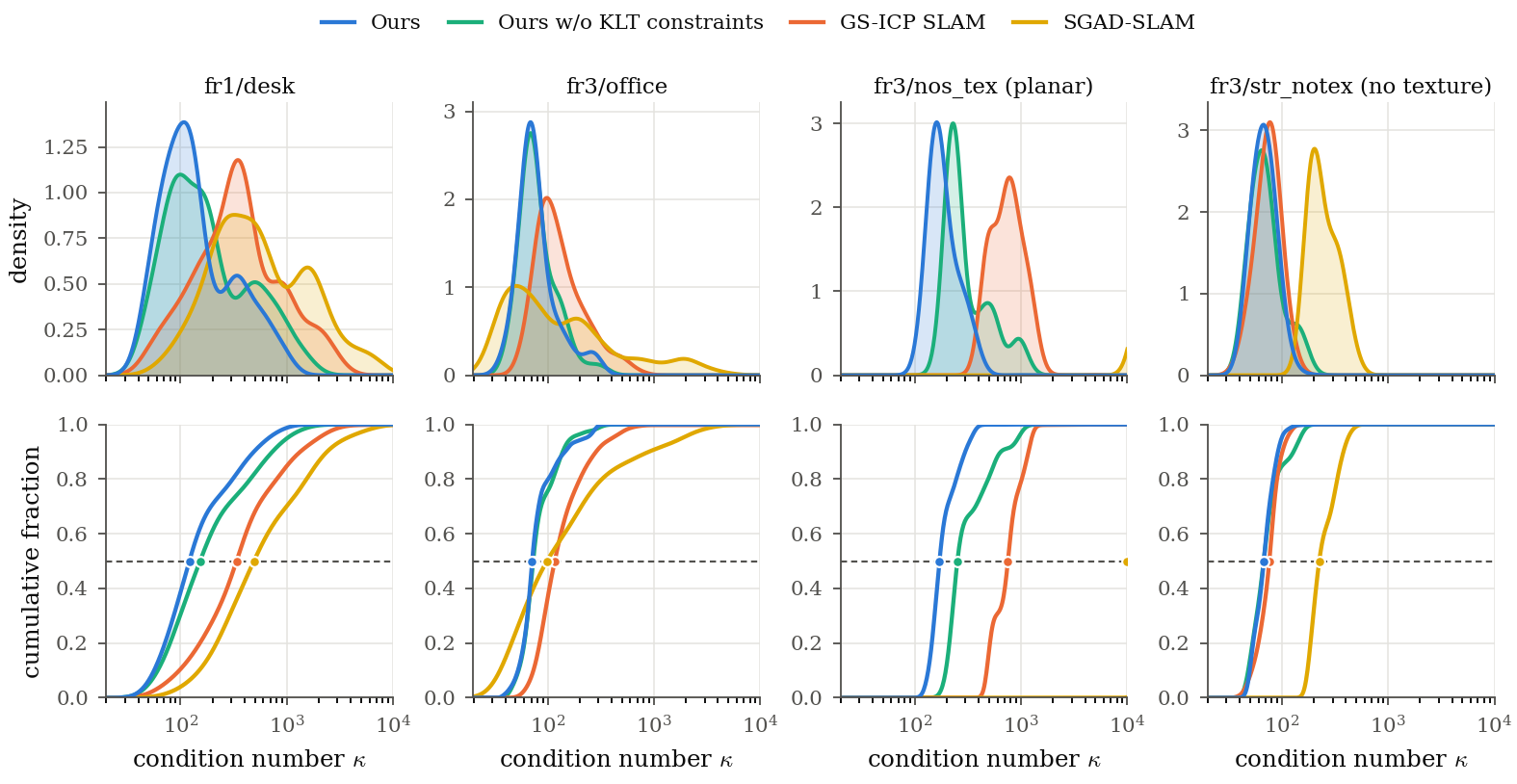}
\caption{\textbf{Comparison of condition number} across ICP methods. Ours reduces the condition number in general, indicating that the optimization is more well-constrained.}\label{fig:k_reduced}
\end{figure}

The proposed covariance is only locally derived from the RGB-D measurement model, whereas pose observability emerges only after aggregating information from all visible primitives. We therefore examine the spectrum of the Hessian \(H=\sum J^\top\Omega^{-1}J\) using its condition number \cite{zhangDegeneracyOptimizationbasedState2016} as a proxy for observability. Replacing the rendering covariance with $\Sigma_t$ changes the uncertainty weighting of individual geometric observations, while the image constraints provide complementary information. Figure~\ref{fig:k_reduced} shows that $\Sigma_t$ substantially improves Hessian conditioning even without image constraints, indicating that the improvement cannot be attributed solely to KLT. In \texttt{str\_notex}, the condition number changes little because the scene already provides strong geometric constraints (from the perpendicular walls and floors). In scenes with weaker geometric constraints, adding KLT constraints further improves the conditioning by contributing complementary image-based information.

\subsection{Ablation studies.} 
In Table~\ref{tab:abl}, we ablate our two main contributions, the tracking covariance $\Sigma_t$ in Section~\ref{subsec:trackcov} and the KLT image residuals in Section~\ref{subsec:imgres}. ``Shared cov" refers to the baseline's method where $\Sigma_t = \Sigma_r$. We show that the image residuals are inconsistently helpful, as fusing them with the shared covariance gives worse tracking in some cases, even with multiple attempted weighting values. On the other hand, $\Sigma_t$ alone improves ATE across real scenes, showing that the representation change does indeed contribute to better tracking. Lastly, our full method achieves the overall lowest ATE, and in particular helps the \texttt{nos\_tex} scene since the scene is otherwise unobservable from depth constraints alone.
\begin{table}[t]\centering
\caption{Ablation of our method, ATE [cm] $\downarrow$.}\label{tab:abl}
\footnotesize\setlength{\tabcolsep}{3.5pt}
\resizebox{\columnwidth}{!}{%
\begin{tabular}{l cccccc}
\toprule
Configuration & fr1/desk & fr2/xyz & fr3/office & fr3/nos\_tex & 0106 & 0169\\
\midrule
Shared cov. & 2.66 & 1.78 & 2.70 & 194.96 & 6.04 & 19.32\\
Shared cov. + KLT ($w$=1) & 3.13 & 1.82 & 11.73 & 4.08 & 5.73 & 20.68 \\
Shared cov. + KLT ($w$=.1) & 2.56 & 1.58 & 4.20 & 3.76 & 8.45 &  9.21 \\
Shared cov. + KLT ($w$=10) & 2.98 & 2.16 & 11.91 & 6.07 & 17.32 & 42.09 \\
$\Sigma_t$ only       & 2.47 & 1.60 & \textbf{2.22} & 192.38 & 5.68 & 14.78\\
Full & \textbf{1.94} & \textbf{1.55} & 2.23 & \textbf{1.67} & \textbf{5.59} & \textbf{6.89}\\
\bottomrule
\end{tabular}%
}
\end{table}

\section{LIMITATIONS}\label{sec:lim}
Although our method is able to robustly track across diverse scene types, limitations exist in terms of large-scale scenes, where our method lacks bundle adjustment to ensure long-term consistency. Like other ICP methods, our work is also sensitive to high depth error, which for many sensors increases with distance. This means that for outdoor scenes or where the majority of the image content is far away, tracking would be less accurate. Future work may consider using the landmarks or covariances for some form of bundle adjustment without trading off too much speed.

\section{CONCLUSION}
We proposed Dual Covariance Gaussian Splatting SLAM, a method in which each Gaussian primitive for mapping carries a second covariance used to store the measurement uncertainty. We showed that our method results in improved tracking performance across the board, including in difficult scenes with fast handheld motion, lack of structure or visual features, and reduced odometry drift. This is done while maintaining real-time performance, enabling deployment on robot platforms for online tracking and mapping.

\bibliographystyle{IEEEtran}
\bibliography{references}

@article{khoshelhamAccuracyResolutionKinect2012,
	title = {Accuracy and {Resolution} of {Kinect} {Depth} {Data} for {Indoor} {Mapping} {Applications}},
	volume = {12},
	copyright = {http://creativecommons.org/licenses/by/3.0/},
	issn = {1424-8220},
	url = {https://www.mdpi.com/1424-8220/12/2/1437},
	doi = {10.3390/s120201437},
	language = {en},
	number = {2},
	urldate = {2026-09-21},
	journal = {Sensors},
	publisher = {Molecular Diversity Preservation International},
	author = {Khoshelham, Kourosh and Elberink, Sander Oude},
	month = feb,
	year = {2012},
	pages = {1437--1454},
}

@inproceedings{schoenberger2016sfm,
	title = {Structure-from-motion revisited},
	booktitle = {Conference on computer vision and pattern recognition ({CVPR})},
	author = {Schönberger, Johannes Lutz and Frahm, Jan-Michael},
	year = {2016},
}

@incollection{henryRGBDMappingUsing2014,
	address = {Berlin, Heidelberg},
	title = {{RGB}-{D} {Mapping}: {Using} {Depth} {Cameras} for {Dense} {3D} {Modeling} of {Indoor} {Environments}},
	isbn = {978-3-642-28572-1},
	shorttitle = {{RGB}-{D} {Mapping}},
	doi = {10.1007/978-3-642-28572-1\_33},
	language = {en},
	urldate = {2026-09-21},
	booktitle = {Experimental {Robotics}: {The} 12th {International} {Symposium} on {Experimental} {Robotics}},
	publisher = {Springer},
	author = {Henry, Peter and Krainin, Michael and Herbst, Evan and Ren, Xiaofeng and Fox, Dieter},
	editor = {Khatib, Oussama and Kumar, Vijay and Sukhatme, Gaurav},
	year = {2014},
	pages = {477--491},
}

@article{whelanElasticFusion2016,
	title = {{ElasticFusion}},
	volume = {35},
	issn = {0278-3649},
	url = {https://doi.org/10.1177/0278364916669237},
	doi = {10.1177/0278364916669237},
	number = {14},
	urldate = {2026-09-21},
	journal = {International Journal of Robotics Research},
	author = {Whelan, Thomas and Salas-Moreno, Renato F and Glocker, Ben and Davison, Andrew J and Leutenegger, Stefan},
	month = dec,
	year = {2016},
	pages = {1697--1716},
}

@misc{yugayGaussianSLAMPhotorealisticDense2024,
	title = {Gaussian-{SLAM}: {Photo}-realistic {Dense} {SLAM} with {Gaussian} {Splatting}},
	shorttitle = {Gaussian-{SLAM}},
	url = {http://arxiv.org/abs/2312.10070},
	doi = {10.48550/arXiv.2312.10070},
	urldate = {2026-09-20},
	publisher = {arXiv},
	author = {Yugay, Vladimir and Li, Yue and Gevers, Theo and Oswald, Martin R.},
	month = mar,
	year = {2024},
	note = {arXiv:2312.10070 [cs.CV]},
}

@inproceedings{nguyenModelingKinectSensor2012,
	title = {Modeling {Kinect} {Sensor} {Noise} for {Improved} {3D} {Reconstruction} and {Tracking}},
	issn = {1550-6185},
	url = {https://ieeexplore.ieee.org/document/6375037},
	doi = {10.1109/3DIMPVT.2012.84},
	urldate = {2026-09-20},
	booktitle = {Visualization \& {Transmission} 2012 {Second} {International} {Conference} on {3D} {Imaging}, {Modeling}, {Processing}},
	author = {Nguyen, Chuong V. and Izadi, Shahram and Lovell, David},
	month = oct,
	year = {2012},
	note = {ISSN: 1550-6185},
	pages = {524--530},
}

@misc{tanSpectralGSSLAMObservabilityAware2026,
	title = {Spectral {GS}-{SLAM}: {Observability}-{Aware}, {Degeneracy}-{Robust} {Tracking} for {Real}-{Time} {3D} {Gaussian} {Splatting} {SLAM}},
	shorttitle = {Spectral {GS}-{SLAM}},
	url = {http://arxiv.org/abs/2606.21258},
	doi = {10.48550/arXiv.2606.21258},
	urldate = {2026-09-20},
	publisher = {arXiv},
	author = {Tan, Edward Beng Wai and Lam, Siew-Kei and Zhang, Dongshuo},
	month = jun,
	year = {2026},
	note = {arXiv:2606.21258 [cs.RO]},
}

@misc{thirgoodFeatureSLAMFeatureenriched3D2026,
	title = {{FeatureSLAM}: {Feature}-enriched {3D} gaussian splatting {SLAM} in real time},
	shorttitle = {{FeatureSLAM}},
	url = {http://arxiv.org/abs/2601.05738},
	doi = {10.48550/arXiv.2601.05738},
	urldate = {2026-09-20},
	publisher = {arXiv},
	author = {Thirgood, Christopher and Mendez, Oscar and Ling, Erin and Storey, Jon and Hadfield, Simon},
	month = mar,
	year = {2026},
	note = {arXiv:2601.05738 [cs.CV]},
}

@inproceedings{Hu2026sgadslam,
	title = {{SGAD}-{SLAM}: {Splatting} gaussians at adjusted depth for better radiance fields in {RGBD} {SLAM}},
	booktitle = {Proceedings of the {IEEE}/{CVF} conference on computer vision and pattern recognition},
	author = {Hu, Pengchong and Han, Zhizhong},
	year = {2026},
}

@inproceedings{daiScanNetRichlyAnnotated3D2017,
	address = {Honolulu, HI},
	title = {{ScanNet}: {Richly}-{Annotated} {3D} {Reconstructions} of {Indoor} {Scenes}},
	isbn = {978-1-5386-0457-1},
	shorttitle = {{ScanNet}},
	url = {https://ieeexplore.ieee.org/document/8099744/},
	doi = {10.1109/CVPR.2017.261},
	urldate = {2026-05-16},
	booktitle = {2017 {IEEE} {Conference} on {Computer} {Vision} and {Pattern} {Recognition} ({CVPR})},
	publisher = {IEEE},
	author = {Dai, Angela and Chang, Angel X. and Savva, Manolis and Halber, Maciej and Funkhouser, Thomas and Niessner, Matthias},
	month = jul,
	year = {2017},
	pages = {2432--2443},
}

@misc{pakG2SICPSLAMGeometryaware2025,
	title = {{G2S}-{ICP} {SLAM}: {Geometry}-aware {Gaussian} {Splatting} {ICP} {SLAM}},
	shorttitle = {{G2S}-{ICP} {SLAM}},
	url = {http://arxiv.org/abs/2507.18344},
	doi = {10.48550/arXiv.2507.18344},
	urldate = {2026-02-25},
	publisher = {arXiv},
	author = {Pak, Gyuhyeon and Cho, Hae Min and Kim, Euntai},
	month = jul,
	year = {2025},
	note = {arXiv:2507.18344 [cs]},
}

@article{straubReplicaDatasetDigital2019,
	title = {The {Replica} {Dataset}: {A} {Digital} {Replica} of {Indoor} {Spaces}},
	journal = {arXiv preprint arXiv:1906.05797},
	author = {Straub, Julian and Whelan, Thomas and Ma, Lingni and Chen, Yufan and Wijmans, Erik and Green, Simon and Engel, Jakob J. and Mur-Artal, Raul and Ren, Carl and Verma, Shobhit and Clarkson, Anton and Yan, Mingfei and Budge, Brian and Yan, Yajie and Pan, Xiaqing and Yon, June and Zou, Yuyang and Leon, Kimberly and Carter, Nigel and Briales, Jesus and Gillingham, Tyler and Mueggler, Elias and Pesqueira, Luis and Savva, Manolis and Batra, Dhruv and Strasdat, Hauke M. and Nardi, Renzo De and Goesele, Michael and Lovegrove, Steven and Newcombe, Richard},
	year = {2019},
}

@inproceedings{segalGeneralizedICP2009,
	title = {Generalized-{ICP}},
	booktitle = {Robotics: {Science} and {Systems}},
	author = {Segal, Aleksandr V. and Hähnel, Dirk and Thrun, Sebastian},
	year = {2009},
}

@inproceedings{zhangDegeneracyOptimizationbasedState2016,
	title = {On degeneracy of optimization-based state estimation problems},
	doi = {10.1109/ICRA.2016.7487211},
	urldate = {2026-02-16},
	booktitle = {2016 {IEEE} {International} {Conference} on {Robotics} and {Automation} ({ICRA})},
	author = {Zhang, Ji and Kaess, Michael and Singh, Sanjiv},
	month = may,
	year = {2016},
	pages = {809--816},
}

@article{kerbl3DGaussianSplatting2023,
	title = {{3D} {Gaussian} {Splatting} for {Real}-{Time} {Radiance} {Field} {Rendering}},
	volume = {42},
	issn = {0730-0301},
	doi = {10.1145/3592433},
	number = {4},
	urldate = {2026-02-19},
	journal = {ACM Trans. Graph.},
	author = {Kerbl, Bernhard and Kopanas, Georgios and Leimkuehler, Thomas and Drettakis, George},
	month = jul,
	year = {2023},
	pages = {139:1--139:14},
}

@article{zhouwangImageQualityAssessment2004,
	title = {Image quality assessment: from error visibility to structural similarity},
	volume = {13},
	copyright = {https://ieeexplore.ieee.org/Xplorehelp/downloads/license-information/IEEE.html},
	issn = {1057-7149, 1941-0042},
	shorttitle = {Image quality assessment},
	doi = {10.1109/TIP.2003.819861},
	number = {4},
	urldate = {2026-02-19},
	journal = {IEEE Transactions on Image Processing},
	author = {{Zhou Wang} and Bovik, A.C. and Sheikh, H.R. and Simoncelli, E.P.},
	month = apr,
	year = {2004},
	pages = {600--612},
}

@misc{yueLPICPGeneralLocalizabilityAware2025,
	title = {{LP}-{ICP}: {General} {Localizability}-{Aware} {Point} {Cloud} {Registration} for {Robust} {Localization} in {Extreme} {Unstructured} {Environments}},
	copyright = {arXiv.org perpetual, non-exclusive license},
	shorttitle = {{LP}-{ICP}},
	doi = {10.48550/ARXIV.2501.02580},
	urldate = {2026-02-19},
	publisher = {arXiv},
	author = {Yue, Haosong and Xu, Qingyuan and Chen, Fei and Pan, Jia and Chen, Weihai},
	year = {2025},
	note = {Version Number: 3},
}

@article{camposORBSLAM3AccurateOpenSource2021,
	title = {{ORB}-{SLAM3}: {An} {Accurate} {Open}-{Source} {Library} for {Visual}, {Visual}–{Inertial}, and {Multimap} {SLAM}},
	volume = {37},
	issn = {1941-0468},
	shorttitle = {{ORB}-{SLAM3}},
	doi = {10.1109/TRO.2021.3075644},
	number = {6},
	urldate = {2026-02-19},
	journal = {IEEE Transactions on Robotics},
	author = {Campos, Carlos and Elvira, Richard and Rodríguez, Juan J. Gómez and M. Montiel, José M. and D. Tardós, Juan},
	month = dec,
	year = {2021},
	pages = {1874--1890},
}

@inproceedings{haRGBDGSICPSLAM2025,
	address = {Cham},
	title = {{RGBD} {GS}-{ICP} {SLAM}},
	volume = {15094},
	isbn = {978-3-031-72763-4 978-3-031-72764-1},
	doi = {10.1007/978-3-031-72764-1_11},
	language = {en},
	urldate = {2026-02-19},
	booktitle = {Computer {Vision} – {ECCV} 2024},
	publisher = {Springer Nature Switzerland},
	author = {Ha, Seongbo and Yeon, Jiung and Yu, Hyeonwoo},
	editor = {Leonardis, Aleš and Ricci, Elisa and Roth, Stefan and Russakovsky, Olga and Sattler, Torsten and Varol, Gül},
	year = {2025},
	pages = {180--197},
}

@article{tunaXICPLocalizabilityAwareLiDAR2024,
	title = {X-{ICP}: {Localizability}-{Aware} {LiDAR} {Registration} for {Robust} {Localization} in {Extreme} {Environments}},
	volume = {40},
	issn = {1941-0468},
	shorttitle = {X-{ICP}},
	doi = {10.1109/TRO.2023.3335691},
	urldate = {2026-02-19},
	journal = {IEEE Transactions on Robotics},
	author = {Tuna, Turcan and Nubert, Julian and Nava, Yoshua and Khattak, Shehryar and Hutter, Marco},
	year = {2024},
	pages = {452--471},
}

@inproceedings{keethaSplaTAMSplatTrack2024,
	title = {{SplaTAM}: {Splat} {Track} \& {Map} {3D} {Gaussians} for {Dense} {RGB}-{D} {SLAM}},
	shorttitle = {{SplaTAM}},
	language = {en},
	urldate = {2025-11-24},
	author = {Keetha, Nikhil and Karhade, Jay and Jatavallabhula, Krishna Murthy and Yang, Gengshan and Scherer, Sebastian and Ramanan, Deva and Luiten, Jonathon},
	year = {2024},
	pages = {21357--21366},
}

@inproceedings{yanGSSLAMDenseVisual2024,
	title = {{GS}-{SLAM}: {Dense} {Visual} {SLAM} with {3D} {Gaussian} {Splatting}},
	shorttitle = {{GS}-{SLAM}},
	language = {en},
	urldate = {2025-11-24},
	author = {Yan, Chi and Qu, Delin and Xu, Dan and Zhao, Bin and Wang, Zhigang and Wang, Dong and Li, Xuelong},
	year = {2024},
	pages = {19595--19604},
}

@inproceedings{matsukiGaussianSplattingSLAM2024,
	title = {Gaussian {Splatting} {SLAM}},
	language = {en},
	urldate = {2025-11-24},
	author = {Matsuki, Hidenobu and Murai, Riku and Kelly, Paul H. J. and Davison, Andrew J.},
	year = {2024},
	pages = {18039--18048},
}

@inproceedings{zhangUnreasonableEffectivenessDeep2018,
	title = {The {Unreasonable} {Effectiveness} of {Deep} {Features} as a {Perceptual} {Metric}},
	booktitle = {Proceedings of the {IEEE} {Conference} on {Computer} {Vision} and {Pattern} {Recognition} ({CVPR})},
	author = {Zhang, Richard and Isola, Phillip and Efros, Alexei A. and Shechtman, Eli and Wang, Oliver},
	month = jun,
	year = {2018},
}

@inproceedings{sturmBenchmarkEvaluationRGBD2012,
	title = {A {Benchmark} for the {Evaluation} of {RGB}-{D} {SLAM} {Systems}},
	booktitle = {Proc. of the {International} {Conference} on {Intelligent} {Robot} {Systems} ({IROS})},
	author = {Sturm, J. and Engelhard, N. and Endres, F. and Burgard, W. and Cremers, D.},
	month = oct,
	year = {2012},
}

@inproceedings{huangPhotoSLAMRealtimeSimultaneous2024,
	title = {Photo-{SLAM}: {Real}-time {Simultaneous} {Localization} and {Photorealistic} {Mapping} for {Monocular} {Stereo} and {RGB}-{D} {Cameras}},
	booktitle = {Proceedings of the {IEEE}/{CVF} {Conference} on {Computer} {Vision} and {Pattern} {Recognition} ({CVPR})},
	author = {Huang, Huajian and Li, Longwei and Cheng, Hui and Yeung, Sai-Kit},
	month = jun,
	year = {2024},
	pages = {21584--21593},
}

\addtolength{\textheight}{-12cm}
\end{document}